\documentclass[12pt]{article}

\usepackage{amsmath}
\usepackage{amssymb}
\usepackage{siunitx}

\usepackage{booktabs}
\usepackage{graphicx}
\usepackage{caption}
\usepackage{subcaption}
\usepackage{multirow}
\usepackage{float}

\usepackage{geometry}
\usepackage{textcomp}

\usepackage[numbers]{natbib}

\usepackage[colorlinks=true,
            linkcolor=black,
            citecolor=black,
            urlcolor=blue]{hyperref}
\usepackage{orcidlink}

\usepackage{setspace}
\usepackage{authblk}

\newcommand{\weik}{w_{\mathrm{eik}}}

\newcommand{\Lpde}{\mathcal{L}_{\mathrm{pde}}}
\newcommand{\Leik}{\mathcal{L}_{\mathrm{eik}}}
\newcommand{\Lic}{\mathcal{L}_{\mathrm{ic}}}

\begin{document}

\title{A Data-Free Physics-Informed Neural Operator for\\
       Level-Set Interface Advection}

\author[1]{Muhammad Akbar Khan~\orcidlink{0009-0001-7956-0080}%
\thanks{Email: \href{mailto:akbar.bsma1337@gmail.com}{akbar.bsma1337@gmail.com}}}
\affil[1]{Department of Mathematics, NED University of Engineering and
          Technology, Karachi 75270, Pakistan}

\date{}

\maketitle

\begin{abstract}
\noindent
Operators for interfacial problems are trained on reference solutions produced
by the solver they are intended to replace. This work develops a data-free 
physics-informed neural operator for level-set
interface advection, in which the interface is the equation's unknown and the
operator maps an initial interface to the full spatiotemporal trajectory under
a prescribed flow. Training uses only the transport residual and a geometric
constraint; no reference solution enters the objective at any point. A
spacetime Fourier backbone emits the entire trajectory in one pass, and the
initial condition is imposed by construction rather than by penalty, which
removes the competition between the anchoring term and the residual that
otherwise arises when no solution data are available. Supervised and hybrid
operators are trained under an identical architecture, family, budget and test
set, and are reported throughout as baselines that quantify what refusing
labels costs. On a reversed single vortex the data-free operator reaches
$1.614 \pm 0.067\%$ relative $L^2$ error on $100$ held-out initial interfaces
against $0.369 \pm 0.035\%$ for the supervised baseline, a factor of $4.4$; on
solid-body rotation the corresponding figures are $3.804 \pm 1.075\%$ and
$2.576 \pm 0.159\%$, a factor of $1.5$. The two benchmarks rank the arms
differently, and the difference is attributable to the eikonal constraint: the
exact solution violates $|\nabla\phi| = 1$ over $0.3\%$ of the domain under
rotation and $86.9\%$ under the vortex. Where the constraint is valid the
physics-trained operator conserves enclosed area $2.7$ times better than the
supervised baseline despite a larger field error, and a hybrid arm using eight
reference solutions outperforms a supervised arm using sixteen.
\end{abstract}

\vspace{2mm}

\vspace{4mm}

\section{Introduction}
\label{sec:intro}

Accurate transport of a moving phase boundary under a prescribed velocity
field is the kinematic core of two-phase flow, solidification and
free-boundary problems. The level-set method~\citep{Osher1988,Osher2003}
represents such a boundary implicitly, as the zero contour of a smooth scalar
field, and thereby handles merging and breakup without special treatment.
Accurate transport requires high-order upwind
discretisation~\citep{shu1988eno,jiang1996weno}, and the convenience of the
implicit representation is offset by a well-known weakness: numerical diffusion
distorts the field over long integration times and the enclosed phase volume
drifts~\citep{Sussman1994}, motivating reinitialisation
procedures~\citep{SussmanFatemi1999} and hybrid particle
corrections~\citep{Enright2002}. Mass conservation remains the principal
weakness of the formulation, and the measure on which competing schemes are
usually separated.

Classical discretisations of the level-set equation solve one initial
configuration at a time. Every new initial interface, or every new velocity
field, requires a complete solve. For design studies, uncertainty
quantification and any other many-query setting this cost is the binding
constraint.

Physics-informed neural networks (PINNs)~\citep{raissi2019pinn} have been
applied to level-set advection with some success. The present author's systematic
study~\citep{khan2026mlst} characterised the eikonal-weight sensitivity,
sampling and encoding choices that govern accuracy across the standard
benchmarks, and a follow-up~\citep{khan2026saw} addressed the adaptive
weighting of the eikonal constraint in three dimensions. Related work has
treated moving interfaces~\citep{mullins2025pinns,Bi2025,Zhai2026},
interface-aware network architectures~\citep{sarma2024ipinns,Chang2025} and
Stefan problems~\citep{wang2021stefan}, and hybrids of a network with particle
level-set tracking~\citep{chen2025neuralpls}. All of these share the limitation
above: the network is fitted to one problem instance, and a new initial
condition means retraining from scratch.

Operator learning addresses exactly this. Rather than approximating one
solution, a neural operator approximates the solution \emph{map}, so a new
configuration costs a single forward pass. DeepONet~\citep{lu2021deeponet} uses
a branch--trunk factorisation; the Fourier neural
operator~\citep{li2021fno} parameterises a kernel integral in the spectral
domain, giving quasi-linear scaling and discretisation invariance. Extensions
handle several input functions or varying
geometry~\citep{jin2022mionet,zhong2025pigano}.

The obvious route to such an operator is supervised: solve the equation many
times, then fit the map. This is what existing interface
operators do. MHNO~\citep{eshaghi2026mhno} decomposes the temporal direction
across multiple heads and is applied to phase-field models, in which the
interface is diffuse and emerges from an energy gradient flow rather than being
transported as a sharp set. IANO~\citep{wang2026iano} conditions on interface
position and topology and performs well in low-data regimes, but requires
labelled trajectories. FBNO~\citep{long2026fbno} maps free-boundary problems
but is restricted by construction to boundaries that deform
diffeomorphically, so it cannot represent merging or splitting, and
\citet{lee2026operator} model rising bubbles from simulation data. The approach
works, but it inherits the cost it was meant to remove: the training set is
produced by exactly the solver the operator is intended to replace.

Data-free operator learning removes that dependency by replacing the empirical
risk with a residual of the governing equation, so training requires no
reference solutions at all. VINO~\citep{eshaghi2025vino} minimises a
variational energy, WINO~\citep{zhu2026wino} a weak form on unfitted domains,
SCLON~\citep{choi2024sclon} a spectral residual and
PINTO~\citep{boya2025pinto} a transformer backbone; a related line develops the
variational formulation and its
analysis~\citep{xu2024variational,qiu2026variationally}.
Between the two extremes sits PINO~\citep{li2024pino}, which combines a data
term with a residual of the governing partial differential equation, and
thereby reduces, without eliminating, the label
requirement.

None of this work addresses level-set advection, and the reason is not
incidental. In the problems these methods target, the PDE unknown is a
displacement, a temperature or a velocity, and accuracy is measured in a field
norm. In level-set advection the quantity of interest is a codimension-one set
that the field norm barely sees: a solution can have a small $L^2$ error and a
badly displaced or destroyed interface. WINO uses a level-set function, but as
a static descriptor of a fixed domain rather than as the evolving unknown.

This paper develops a data-free physics-informed neural operator for level-set
interface advection. The operator maps an initial interface to the full spatiotemporal solution
trajectory under a prescribed flow, and it is trained entirely on the transport
residual and a geometric constraint. No reference solution enters training at
any point. A supervised operator and a hybrid arm are trained alongside it,
under an identical architecture, family, budget and test set, and are reported
as baselines throughout. Both benchmarks admit exact solutions, so the price of
refusing labels is measured rather than asserted; what that price buys is taken
up in Section~\ref{sec:discussion}.

The specific contributions are:
\begin{enumerate}
  \item A spacetime neural operator for level-set advection trained without
        reference solutions, mapping an initial interface $\phi_0$ to the
        full solution trajectory $\phi(\cdot,t)$ over a family of initial
        configurations, with the initial condition enforced by construction
        rather than by penalty.
  \item A quantitative account of when the eikonal constraint helps and when it
        does not. On rigid rotation the exact solution remains a signed
        distance function and $0.3\%$ of the domain violates
        $|\nabla\phi|=1$; on the reversed vortex $86.9\%$ does. The constraint
        is therefore benchmark-dependent in a way that no single weight
        resolves, and the adaptive scheme of~\citet{khan2026saw} requires
        modification under a hard initial condition.
  \item A measured comparison against supervised and hybrid baselines under
        an identical architecture, family, optimisation budget and test set,
        on two benchmarks chosen at opposite ends of that axis, reporting
        interface conservation alongside field error.
\end{enumerate}

The remainder of the paper is organised as follows.
Section~\ref{sec:levelset} sets out the level-set formulation and the
signed-distance property on which the geometric constraint rests.
Section~\ref{sec:method} describes the operator, the structural initial
condition and the residual objective, and Section~\ref{sec:setup} the
benchmarks, families and error measures. Section~\ref{sec:results} reports the
results and the ablations, Section~\ref{sec:discussion} discusses what they
imply about the geometric constraint, and Section~\ref{sec:conclusion}
concludes.

\section{The Level-Set Method}
\label{sec:levelset}

\subsection{Formulation}

In the level-set framework~\citep{Osher1988}, a moving interface $\Gamma(t)$ is
represented implicitly as the zero contour of a scalar field
$\phi(\mathbf{x},t)$ on a fixed spatial domain
$\Omega \subset \mathbb{R}^2$,
\begin{equation}
  \Gamma(t) = \{\mathbf{x}\in\Omega : \phi(\mathbf{x},t) = 0\},
  \label{eq:interface}
\end{equation}
with the sign of $\phi$ distinguishing the two phases. This work considers pure
kinematic advection under a prescribed velocity field
$\mathbf{u}(\mathbf{x},t)$, and takes $\Omega = [0,1]^2$.

The evolution of $\phi$ is governed by the level-set advection equation
\begin{equation}
  \frac{\partial \phi}{\partial t} + \mathbf{u}\cdot\nabla\phi = 0,
  \qquad \mathbf{x}\in\Omega, \quad t\in[0,T].
  \label{eq:transport}
\end{equation}
Its exact solution is the initial field composed with the backward flow map:
the value at $(\mathbf{x},t)$ is $\phi_0$ evaluated at the foot of the
characteristic through that point. This is the property used in
Section~\ref{sec:benchmarks} to obtain reference solutions where no closed form
exists.

\subsection{Signed-distance regularity}
\label{sec:sdf}

Level-set fields are conventionally initialised as signed distance functions,
satisfying $|\nabla\phi| = 1$, since interface normals and curvature are then
recovered directly from the field gradients. The property is not preserved by
\eqref{eq:transport}: it survives rigid motion, but under deformation the
transported field departs from a distance function even when the transport
itself is exact. Classical schemes restore the condition by
reinitialisation~\citep{Sussman1994,SussmanFatemi1999}, which repositions the
field without moving the zero contour.

Neural formulations instead impose the condition through an eikonal penalty in
the training loss. Its weight is the most sensitive hyperparameter in the
level-set PINN setting, spanning four orders of magnitude between rigid and
deforming benchmarks~\citep{khan2026mlst}. Section~\ref{sec:eikonal} treats the
same term in the operator setting, and Section~\ref{sec:benchmarks} quantifies
how far the exact solution of each benchmark departs from a distance function.

\section{Method}
\label{sec:method}

\subsection{Problem statement}

\begin{figure}[H]
\centering
\includegraphics[width=\textwidth]{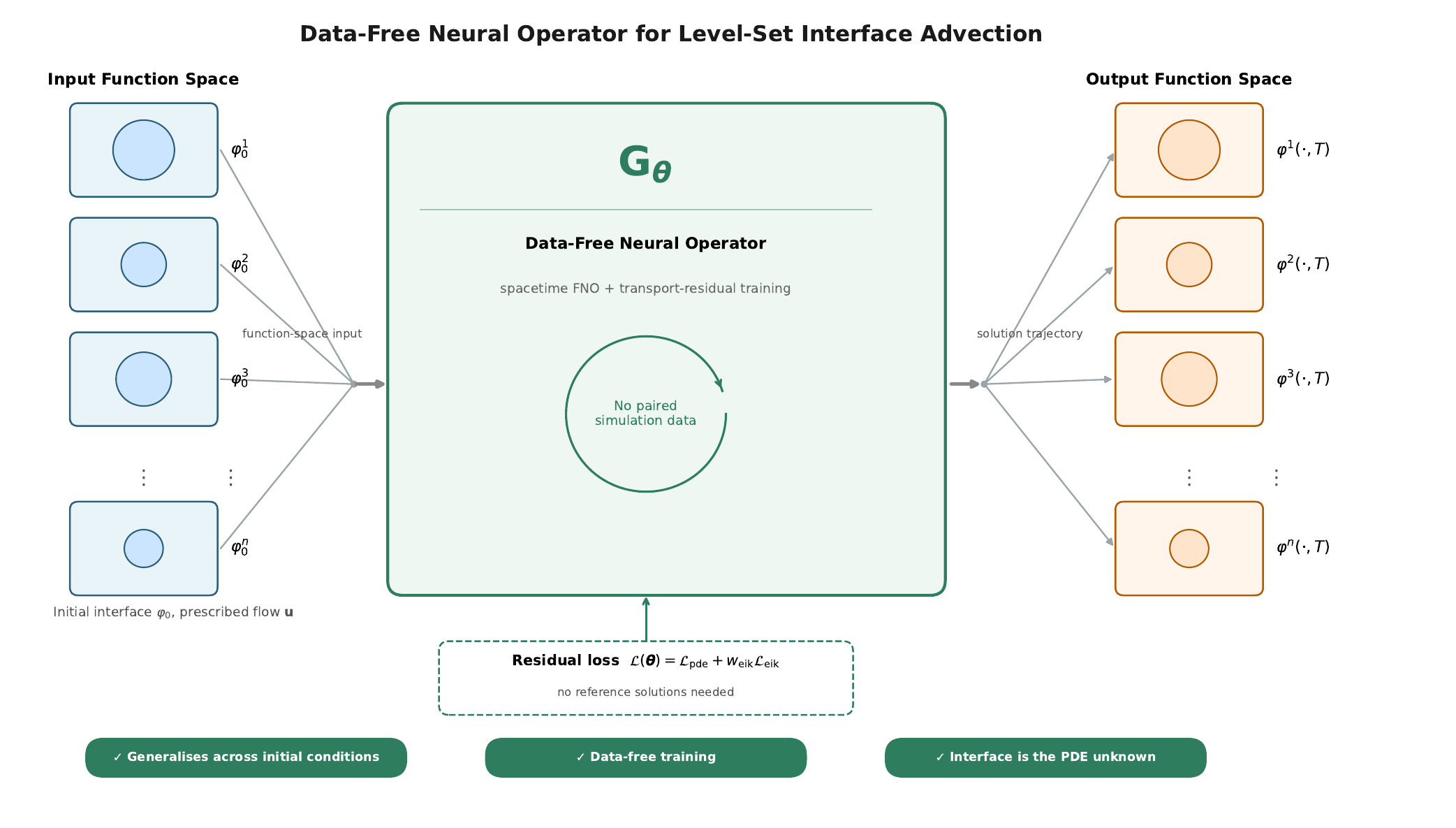}
\caption{The operator maps an initial interface to the full spatiotemporal
trajectory under a prescribed flow. Its parameters are determined from the
transport residual and the eikonal term alone; no reference solution enters
training.}
\label{fig:schematic}
\end{figure}

Let $\Omega = [0,1]^2$ and $T>0$, and let
$\mathbf{u}:\Omega\times[0,T]\to\mathbb{R}^2$ be a prescribed divergence-free
velocity field. Given an initial level-set field $\phi_0:\Omega\to\mathbb{R}$
whose zero contour is the initial interface, the solution operator
\begin{equation}
  \mathcal{G} : \phi_0 \;\longmapsto\; \phi(\cdot,\cdot)
  \label{eq:operator}
\end{equation}
returns the field satisfying~\eqref{eq:transport} with $\phi(\cdot,0)=\phi_0$.
The aim is a parametric approximation $\mathcal{G}_\theta \approx \mathcal{G}$
whose parameters are determined without evaluating $\mathcal{G}$ at any
training input. Figure~\ref{fig:schematic} summarises the construction.

\subsection{Spacetime Fourier neural operator}

$\mathcal{G}_\theta$ is a three-dimensional Fourier neural operator
(FNO)~\citep{li2021fno} over $(x,y,t)$. It takes six input channels:
$\phi_0$ broadcast along time, the two velocity
components, and the three coordinates. The velocity channels are constant
across the family in the experiments reported here, and are retained so that
the same architecture admits a varying flow. These are lifted to a latent
width,
passed through four Fourier layers, and projected back to a scalar field on the
full spacetime grid.

The operator emits the entire trajectory in one forward pass rather than
stepping in time. This is a deliberate choice with two consequences. The
residual is evaluated over $\Omega\times[0,T]$ simultaneously, so no
backpropagation through a rollout is required; and there is no autoregressive
error accumulation, which is the mechanism that destabilises compact FNO
surrogates over long horizons and motivates the multi-head decomposition
of~\citet{eshaghi2026mhno}.

\subsection{Structural enforcement of the initial condition}

The initial condition is imposed by construction,
\begin{equation}
  \mathcal{G}_\theta[\phi_0](\mathbf{x},t)
    = \phi_0(\mathbf{x}) + \frac{t}{T}\,\mathcal{N}_\theta[\phi_0](\mathbf{x},t),
  \label{eq:hardic}
\end{equation}
so that $\mathcal{G}_\theta[\phi_0](\cdot,0) = \phi_0$ exactly for
every input, and no initial-condition penalty appears in the loss. Ans\"atze of
this kind, composing the network output with a function that vanishes where the
condition is prescribed, date to \citet{lagaris1998ann} and have since been
developed systematically for boundary conditions on complex
geometries~\citep{sukumar2022exact} and for inverse design~\citep{lu2021hard}.
The temporal form used here is the standard one: a monotonic factor transfers
weight from the prescribed initial state to the learned correction as time
advances.

The motivation is specific to data-free training. Equation~\eqref{eq:transport}
is satisfied by the transport of \emph{any} initial field; the initial
condition is the only information that selects the intended solution. Imposed
as a soft penalty it competes with the residual, and the optimiser can reduce
the total loss by drifting away from the prescribed initial data. Under
\eqref{eq:hardic} that failure mode is structurally excluded, and the residual
is left with a single objective.

It also changes how the adaptive eikonal weight of Section~\ref{sec:eikonal}
behaves at the first iteration, since the field then begins at an exact
distance function.

\subsection{Residual loss}

The transport residual is evaluated by second-order central differences on the
spacetime grid,
\begin{equation}
  \mathcal{R}[\phi]_{ijk} =
    \left(\partial_t\phi + \mathbf{u}\cdot\nabla\phi\right)_{ijk},
  \qquad
  \mathcal{L}_{\text{pde}} = \frac{1}{|I|}\sum_{ijk \in I} \mathcal{R}[\phi]_{ijk}^2 ,
  \label{eq:pde}
\end{equation}
over interior nodes $I$.

\subsection{The eikonal constraint}
\label{sec:eikonal}

Level-set methods conventionally maintain $|\nabla\phi| = 1$, and the
corresponding penalty
\begin{equation}
  \mathcal{L}_{\text{eik}}
    = \frac{1}{|I|}\sum_{ijk \in I}\bigl(|\nabla\phi|_{ijk} - 1\bigr)^2
  \label{eq:eik}
\end{equation}
is standard in level-set PINNs.

The constraint is not uniformly appropriate. The exact solution
of~\eqref{eq:transport} is the initial field composed with the backward flow
map, and this preserves the signed-distance property only for rigid motion.
Under deformation the true solution violates $|\nabla\phi|=1$, so the penalty
drives the approximation away from the solution it is meant to regularise.
Section~\ref{sec:benchmarks} quantifies this for the two benchmarks used here,
and the difference between them is large: $0.3\%$ of the domain against
$86.9\%$.

SDF-Aware Weighting~\citep{khan2026saw} was introduced for precisely this
situation. It gates the pointwise residuals above a running quantile
$\hat\tau$ before scaling the survivors so that their contribution to the
parameter update matches that of the other terms,
\begin{equation}
  \weik = \operatorname{clip}\!\left(
    \frac{\|\nabla_\theta\Lpde\| + \|\nabla_\theta\Lic\|}
         {\|\nabla_\theta\Leik\| + \varepsilon},\,0,\,w_{\mathrm{pde}}\right),
  \label{eq:saw}
\end{equation}
where $w_{\mathrm{pde}}$ is the upper bound on the eikonal weight, smoothed by
an exponential moving average with decay $\beta$. The gate removes
points at which a large eikonal residual is evidence of genuine departure from
a distance function rather than of network error, and it is the gate that
allows the ratio to settle rather than saturate.

\paragraph{Extension to a structurally imposed initial condition.}
The scheme was developed for a soft initial condition, where an
initial-condition loss $\Lic$ is present and the network is randomly
initialised. Both properties enter~\eqref{eq:saw}. The numerator contains
$\|\nabla_\theta\Lic\|$, and at the first iteration a random network bears no
resemblance to a distance function, so $\|\nabla_\theta\Leik\|$ is large and the
ratio is immediately informative.

Neither property holds under~\eqref{eq:hardic}. There is no $\Lic$ to
differentiate, so the numerator reduces to $\|\nabla_\theta\Lpde\|$; and the
field begins at exactly $\phi_0$, which \emph{is} a distance function, so the
eikonal residual at initialisation is of order $10^{-4}$ and the denominator is
near zero. The ratio is therefore uninformative at the first iteration
specifically, and seeding the moving average with it holds the weight near the
clamp for $O((1-\beta)^{-1})$ subsequent iterations.

The adaptation required is confined to that seeding. Initialising the moving
average at zero rather than at the first ratio leaves the gate, the quantile,
the ratio and the clamp exactly as published, and the weight then rises as the
field departs from $\phi_0$ and the ratio becomes meaningful. The asymmetry is
what makes this the right direction: a weight that is too small early acts when
the field has barely moved and the eikonal term carries little information,
whereas a weight that is too large early can drive the solution into a
configuration from which the later correction does not recover. That a single
change to the initialisation suffices, with the mechanism otherwise untouched,
indicates that the scheme transfers to the operator setting rather than
requiring reformulation for it.

\subsection{Training objective}

The data-free objective is
\begin{equation}
  \mathcal{L} = \mathcal{L}_{\text{pde}} + w_{\text{eik}}\,\mathcal{L}_{\text{eik}} ,
  \label{eq:loss}
\end{equation}
with $w_{\text{eik}}$ set by~\eqref{eq:saw}. No reference solution enters
\eqref{eq:loss}.

\subsection{Baselines}
\label{sec:baselines}

Two baselines are trained under an architecture, family, optimisation budget
and test set identical to those used for~\eqref{eq:loss}. They measure what
training without reference solutions costs.

The \emph{supervised} baseline discards~\eqref{eq:loss} entirely and fits the
exact solution on every training instance,
\begin{equation}
  \mathcal{L}_{\text{sup}} = \frac{1}{|F|\,|I|}\sum_{s\in F}\sum_{ijk\in I}
      \Bigl(\mathcal{G}_\theta\bigl[\phi_0^{(s)}\bigr]_{ijk}
            - \phi^{(s)}_{ijk}\Bigr)^{2} ,
  \label{eq:sup}
\end{equation}
where $F$ indexes the training family and $\phi^{(s)}$ is the exact solution
for instance $s$, evaluated on the same grid nodes $I$ as the
residual~\eqref{eq:pde}. No residual appears: this arm has no knowledge of the
governing equation beyond what the solutions themselves carry.

The \emph{hybrid} baseline, following PINO~\citep{li2024pino},
retains~\eqref{eq:loss} on every instance and adds the same data term on a
labelled subset only,
\begin{equation}
  \mathcal{L}_{\text{hyb}} = \mathcal{L}
    + \frac{\lambda_{\text{data}}}{|S|\,|I|}\sum_{s\in S}\sum_{ijk\in I}
      \Bigl(\mathcal{G}_\theta\bigl[\phi_0^{(s)}\bigr]_{ijk}
            - \phi^{(s)}_{ijk}\Bigr)^{2} ,
  \label{eq:hybrid}
\end{equation}
where $S \subset F$ indexes half the training family. The hybrid arm therefore
interpolates between the two: it carries the residual everywhere and solution
data on half the instances.

\section{Benchmarks and Experimental Setup}
\label{sec:setup}

\subsection{Benchmarks}
\label{sec:benchmarks}

Two benchmarks are used, chosen because they sit at opposite ends of the
deformation range and because exact solutions are available for both.

\paragraph{Solid-body rotation (RO).}
The velocity field is a rigid rotation about the domain centre,
\begin{equation}
  u_1 = -\omega\,(y - \tfrac{1}{2}), \qquad
  u_2 = \phantom{-}\omega\,(x - \tfrac{1}{2}),
  \label{eq:ro}
\end{equation}
with $\omega = 1$ and $T = 2\pi$, so the interface completes one revolution and
returns to its starting position. The initial interface is a circle of radius
$R$ centred at distance $\rho$ from the domain centre at phase $\alpha$, and the
exact solution is that circle rotated rigidly through an angle $\omega t$. Since
the motion is rigid the exact solution remains a signed distance function for
all time.

\paragraph{Reversed single vortex (RV).}
The velocity field is
\begin{equation}
  u_1 = -\sin^2(\pi x)\sin(2\pi y)\cos(\pi t/T), \qquad
  u_2 = \phantom{-}\sin^2(\pi y)\sin(2\pi x)\cos(\pi t/T),
  \label{eq:rv}
\end{equation}
with $T = 2$. The cosine factor reverses the flow at $t = T/2$, so the interface
stretches into a thin filament and then returns to its initial shape; the final
error is magnified by the reversal rather than cancelling with it. The initial
interface is a circle of radius $R$ centred at $(c_x, c_y)$. No closed form
exists, so reference solutions are obtained by backward characteristics:
each evaluation point is integrated back to $t=0$ with RK45 at relative
tolerance $10^{-8}$ and absolute tolerance $10^{-10}$, and $\phi_0$ is evaluated
at the foot of the characteristic. Because the velocity does not depend on the
family member, these back-traced coordinates are computed once and reused by
every instance. Both fields are divergence-free to machine precision.

\paragraph{Departure from the signed-distance property.}
Table~\ref{tab:eikstat} quantifies the difference between the two benchmarks
that motivates Section~\ref{sec:eikonal}. Under rotation the eikonal residual
of the \emph{exact} solution is negligible; under the vortex it is large over
most of the domain.

\begin{table}[H]
\centering
\caption{Eikonal residual $(|\nabla\phi|-1)^2$ of the exact solution,
evaluated on the $64\times64\times32$ grid.}
\label{tab:eikstat}
\begin{tabular}{lcc}
\toprule
& \textbf{median} & \textbf{fraction $>10^{-3}$} \\
\midrule
RO & $2.4\times10^{-8}$ & $0.3\%$ \\
RV & $8.8\times10^{-2}$ & $86.9\%$ \\
\bottomrule
\end{tabular}
\end{table}

\subsection{Input families}

In both benchmarks the velocity field is held fixed and the initial interface
varies, which is the single-input-function protocol used throughout the
operator literature.

Sixteen instances are drawn by Latin hypercube for training and a further
hundred, never seen during training, for evaluation. Table~\ref{tab:family}
gives the ranges, chosen so that the interface stays clear of $\partial\Omega$
for the whole of $[0,T]$.

\begin{table}[H]
\centering
\caption{Family parameters. For RO the circle of radius $R$ is centred at
distance $\rho$ from $(\tfrac{1}{2},\tfrac{1}{2})$ at phase $\alpha$; for RV it
is centred at $(c_x,c_y)$.}
\label{tab:family}
\begin{tabular}{llc}
\toprule
& \textbf{Parameter} & \textbf{Range} \\
\midrule
\multirow{3}{*}{\textbf{RO}}
 & orbit radius $\rho$  & $[0.18,\ 0.26]$ \\
 & phase $\alpha$       & $[0,\ 2\pi)$ \\
 & circle radius $R$    & $[0.10,\ 0.18]$ \\
\midrule
\multirow{3}{*}{\textbf{RV}}
 & centre $c_x$         & $[0.40,\ 0.60]$ \\
 & centre $c_y$         & $[0.68,\ 0.82]$ \\
 & circle radius $R$    & $[0.10,\ 0.18]$ \\
\bottomrule
\end{tabular}
\end{table}

Test instances are drawn from the same distribution and never appear in
training; all reported errors are over $100$ held-out instances.

\subsection{Metrics}

Following~\citet{khan2026mlst}, the relative field error at time $t$ is
\begin{equation}
  \mathcal{E}^{\text{rel}}_{L^2}(t) = 100 \times
    \frac{\|\hat\phi(\cdot,t) - \phi^{\text{ref}}(\cdot,t)\|_2}
         {\|\phi^{\text{ref}}(\cdot,t)\|_2}\ \%,
  \label{eq:rell2}
\end{equation}
averaged over time and over the test set.

Field norms alone are insufficient for interface problems, since the zero
contour occupies a small fraction of the domain and a large field error can
coexist with a well-placed interface, or the reverse. Two area-based measures
are therefore reported, both built on the enclosed area
\begin{equation}
  A[\phi](t) = h_x h_y \sum_{ij} \mathbf{1}\bigl[\phi_{ij}(t) < 0\bigr],
  \label{eq:area}
\end{equation}
the cell count of the region the interface encloses. The first measure compares
this against the reference field,
\begin{equation}
  \mathcal{E}^{\text{area}}(t) = 100 \times
    \frac{\bigl|A[\hat\phi](t) - A[\phi^{\text{ref}}](t)\bigr|}
         {A[\phi^{\text{ref}}](t)}\ \%,
  \label{eq:mass}
\end{equation}
counted identically on both sides so that boundary discretisation cancels. The
second measures drift of the enclosed area from the prediction's own value at
$t=0$,
\begin{equation}
  \mathcal{E}^{\text{drift}}(t) = 100 \times
    \frac{\bigl|A[\hat\phi](t) - A[\hat\phi](0)\bigr|}{A[\hat\phi](0)}\ \%,
  \label{eq:drift}
\end{equation}
and requires no reference at all: both velocity fields are divergence-free, so
the exact area is conserved and any drift is error. Equation~\eqref{eq:drift}
certifies conservation rather than correctness, since a field that never moves
scores zero, and it is reported alongside~\eqref{eq:mass} rather than in place
of it. Both are averaged over time and over the test set, and both are reported
in Tables~\ref{tab:ro-a} and~\ref{tab:rv-a}. They agree to within $0.4$
percentage points on every arm of both benchmarks, so the reference-free
measure is a reliable substitute where no exact solution is available.

The reference-free measure is read against its discretisation floor: on the
$64\times64\times32$ grid the exact solution itself registers $0.82\%$ (RO) and
$1.09\%$ (RV) on the drift measure, and no approximation can report better.

\subsection{Configuration}

All experiments use a $64\times64$ spatial grid with $32$ time levels, a
latent width of $20$ and $(12,12,8)$ retained Fourier modes, giving
$7.4\times10^{6}$ parameters. Training uses Adam~\citep{Kingma2015} for
$30{,}000$ iterations on the rotation benchmark and $20{,}000$ on the vortex,
at batch size four with cosine
annealing~\citep{Loshchilov2017} from $10^{-3}$ to $10^{-5}$, and gradient-norm
clipping at unity. Every arm receives an identical budget: the supervised
baseline converges considerably earlier, but a shorter allocation would render
the comparison unfair in a manner that is difficult to defend in either
direction. The two benchmarks differ because the vortex reaches its final
values by $20{,}000$ iterations and rotation does not;
Section~\ref{sec:res:budget} reports both budgets for rotation.

Each configuration is run with three seeds ($42$, $43$, $44$) and reported as
mean and standard deviation. The hybrid arm uses
$\lambda_{\text{data}} = 1$ on eight of the sixteen training instances, and the
adaptive weighting uses $\beta = 0.999$, $q = 0.95$ and $w_{\mathrm{pde}} = 1$.
All computation was performed on a single NVIDIA Tesla T4.

\section{Results}
\label{sec:results}

Results are reported benchmark by benchmark. Within each, the three arms share
an identical architecture, family, optimisation budget and test set, so the
only difference between them is the loss. Every number is the mean over three
seeds with the sample standard deviation, evaluated on $100$ held-out
instances.

\subsection{Solid-body rotation, varying initial condition}
\label{sec:res:ro:a}

Table~\ref{tab:ro-a} reports the rigid benchmark. Unlike the deforming case
treated next, the ordering here is not monotone in the number of reference
solutions, and the two error measures do not agree on a ranking.

\begin{table}[H]
\centering
\caption{Solid-body rotation, varying initial condition. Sixteen training
instances, $30{,}000$ iterations, three seeds, $100$ held-out instances.
Area error is measured against the reference field~\eqref{eq:mass} and drift
against the prediction's own initial area~\eqref{eq:drift}; the latter requires
no reference and its discretisation floor is given in the last row.}
\label{tab:ro-a}
\begin{tabular}{lcccc}
\toprule
\textbf{Arm} & \textbf{Labels} & \textbf{Relative $L^2$ (\%)} &
\textbf{Area error (\%)} & \textbf{Drift (\%)} \\
\midrule
Data-free   & $0$  & $3.804 \pm 1.075$ & $\mathbf{3.15 \pm 0.49}$ & $3.28 \pm 0.47$ \\
Hybrid      & $8$  & $\mathbf{1.599 \pm 0.084}$ & $3.42 \pm 0.28$ & $3.54 \pm 0.29$ \\
Supervised  & $16$ & $2.576 \pm 0.159$ & $8.61 \pm 0.97$ & $8.91 \pm 0.99$ \\
\midrule
Exact solution & --- & $0$ & $0$ & $0.82$ \\
\bottomrule
\end{tabular}
\end{table}

\paragraph{Eight labels with the residual outperform sixteen without it.}
The hybrid arm reaches $1.599\%$ against $2.576\%$ for the supervised arm,
using half as many reference solutions. Adding the remaining eight labels and
removing the residual makes the result worse. This does not occur on the
deforming benchmark of Section~\ref{sec:res:rv:a}, where the ordering is
monotone, and the difference between the two cases is examined below.

\begin{figure}[H]
\centering
\includegraphics[width=\textwidth]{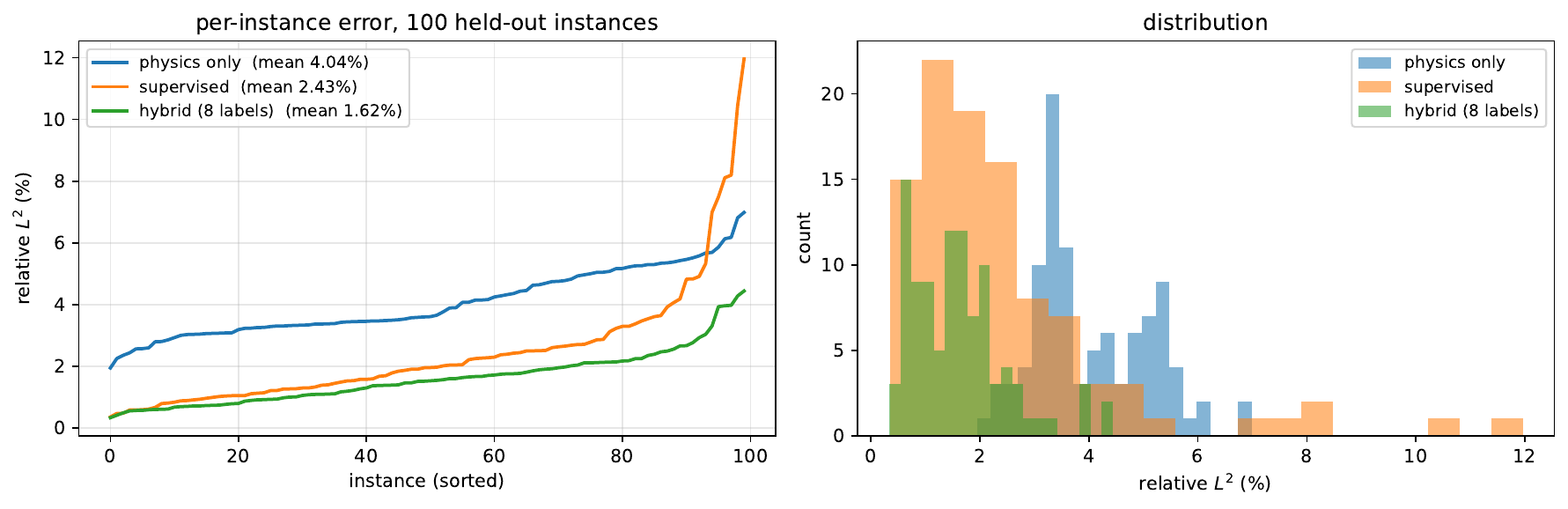}
\caption{Solid-body rotation: relative $L^2$ error on each of the $100$
held-out instances, sorted (left) and as a distribution (right), seed $42$.}
\label{fig:ro-spread}
\end{figure}

\paragraph{The data-free arm conserves area best.}
At $3.15\%$ it is better than the supervised arm by a factor of $2.7$, despite
a relative $L^2$ error $1.5$ times larger. The supervised arm also has the
longer tail: Figure~\ref{fig:ro-spread} shows its worst held-out instance at
almost twice the error of the worst data-free one. The two arms therefore do
not fail in the same way, and the error fields of Figure~\ref{fig:ro-fields}
show the difference: the supervised
residual is high-frequency and distributed across the domain including the
interface band, whereas the data-free residual is smooth, concentrated away
from the zero contour, and structured along the medial axis where the distance
function is not differentiable.

\paragraph{The eikonal constraint is active on this benchmark, and its weight
is seed-dependent.}
The exact solution here remains a signed distance function
(Table~\ref{tab:eikstat}), so the geometric term supplies genuine information
rather than competing with the transport residual. The adaptive weight
consequently settles at $O(10^{-3})$ rather than the $O(10^{-4})$ reached on
the vortex, but the value it selects varies across seeds by a factor of
eleven, and the resulting errors are ordered identically to those weights. The
data-free standard deviation of $1.075$ percentage points, an order of magnitude
above the other two arms, is attributable to this. It is a
property of the adaptive scheme on a benchmark where the constraint is active,
not an artefact of incomplete training: raising the budget from $20{,}000$ to
$30{,}000$ iterations improved the mean by $0.9$ percentage points while
reducing the spread by only $0.2$ (Section~\ref{sec:res:budget}).

\begin{figure}[H]
\centering
\includegraphics[width=\textwidth]{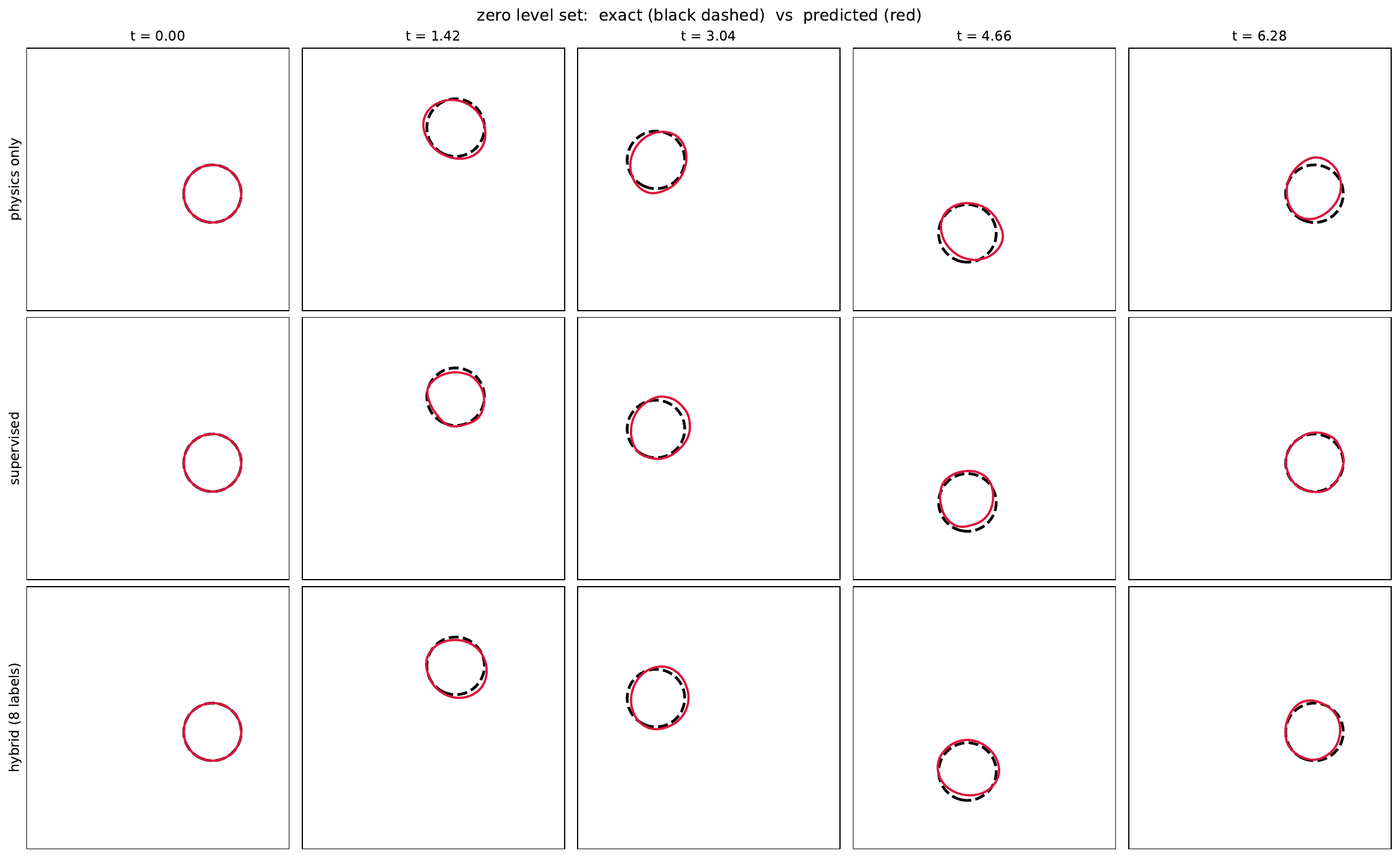}
\caption{Solid-body rotation: predicted zero level set (solid) against the
exact interface (dashed) at five times, for one held-out initial interface.
Rows are the three arms at seed $42$.}
\label{fig:ro-interfaces}
\end{figure}

\begin{figure}[H]
\centering
\includegraphics[width=\textwidth]{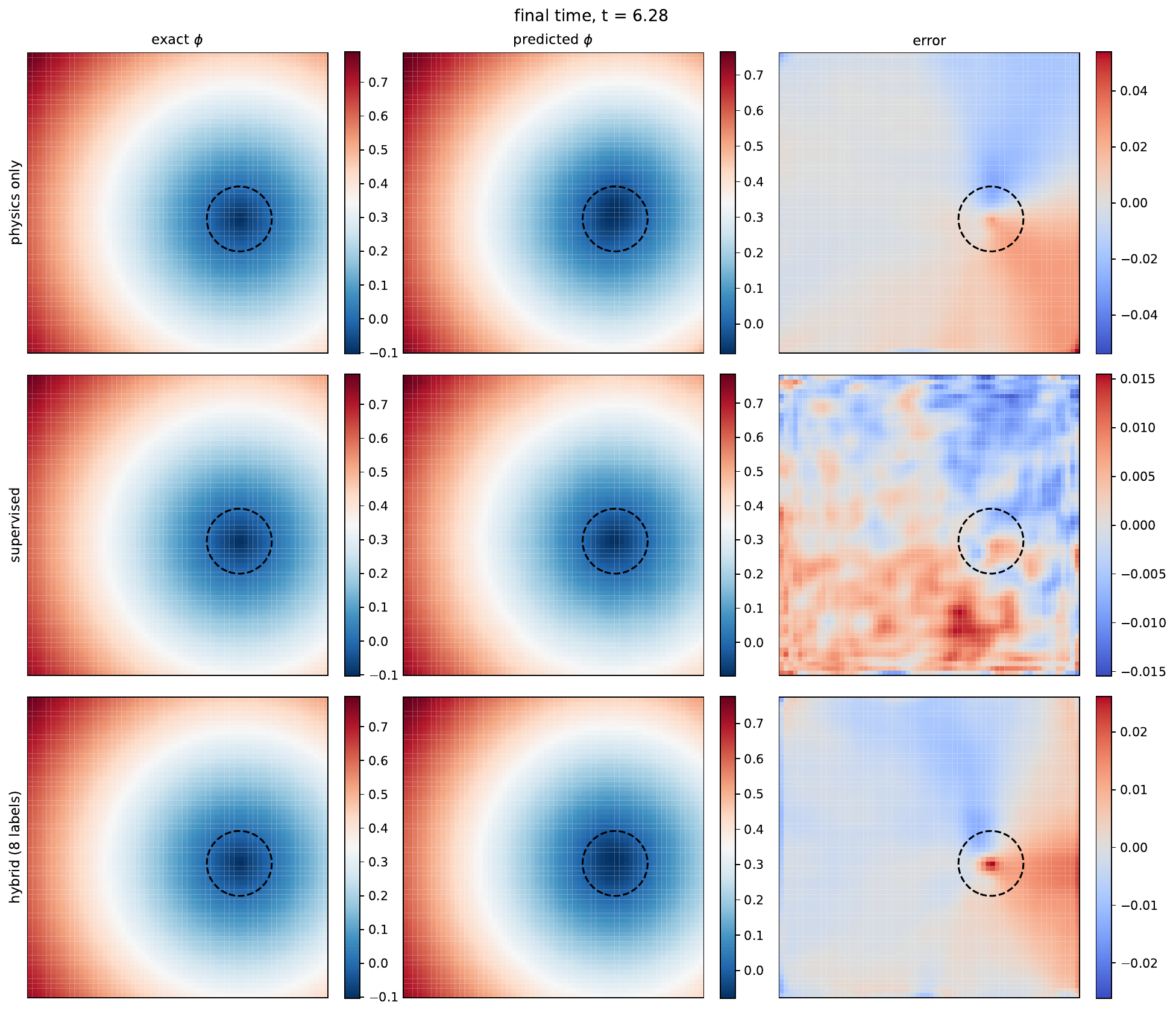}
\caption{Solid-body rotation at the final time: exact field, prediction and
signed error for the same held-out instance as
Figure~\ref{fig:ro-interfaces}. The dashed contour marks the exact interface.
Note the differing colour scales.}
\label{fig:ro-fields}
\end{figure}

\subsection{Reversed vortex, varying initial condition}
\label{sec:res:rv:a}

Table~\ref{tab:rv-a} reports the deforming benchmark with the velocity field
held fixed. The ordering is monotone in the number of reference solutions used,
on both the field error and the enclosed-area error.

\begin{table}[H]
\centering
\caption{Reversed vortex, varying initial condition. Sixteen training
instances, $20{,}000$ iterations, three seeds, $100$ held-out instances.
Columns as in Table~\ref{tab:ro-a}.}
\label{tab:rv-a}
\begin{tabular}{lcccc}
\toprule
\textbf{Arm} & \textbf{Labels} & \textbf{Relative $L^2$ (\%)} &
\textbf{Area error (\%)} & \textbf{Drift (\%)} \\
\midrule
Data-free   & $0$  & $1.614 \pm 0.067$ & $3.57 \pm 0.79$ & $3.68 \pm 0.78$ \\
Hybrid      & $8$  & $0.986 \pm 0.081$ & $2.73 \pm 0.56$ & $2.83 \pm 0.60$ \\
Supervised  & $16$ & $\mathbf{0.369 \pm 0.035}$ & $\mathbf{1.13 \pm 0.10}$ & $\mathbf{1.50 \pm 0.13}$ \\
\midrule
Exact solution & --- & $0$ & $0$ & $1.09$ \\
\bottomrule
\end{tabular}
\end{table}

Three observations follow.

\paragraph{The cost of refusing labels is a factor of $4.4$.}
The operator trained on the transport residual alone reaches $1.614\%$ relative
error on initial interfaces it has never seen, against $0.369\%$ for the same
architecture trained on sixteen exact solutions. No reference solution enters
the data-free objective at any point, so this factor is the price of the
method rather than of its implementation. The hybrid arm sits between the two
at half the label count, which is the behaviour reported for
PINO~\citep{li2024pino} on other equations.

\begin{figure}[H]
\centering
\includegraphics[width=\textwidth]{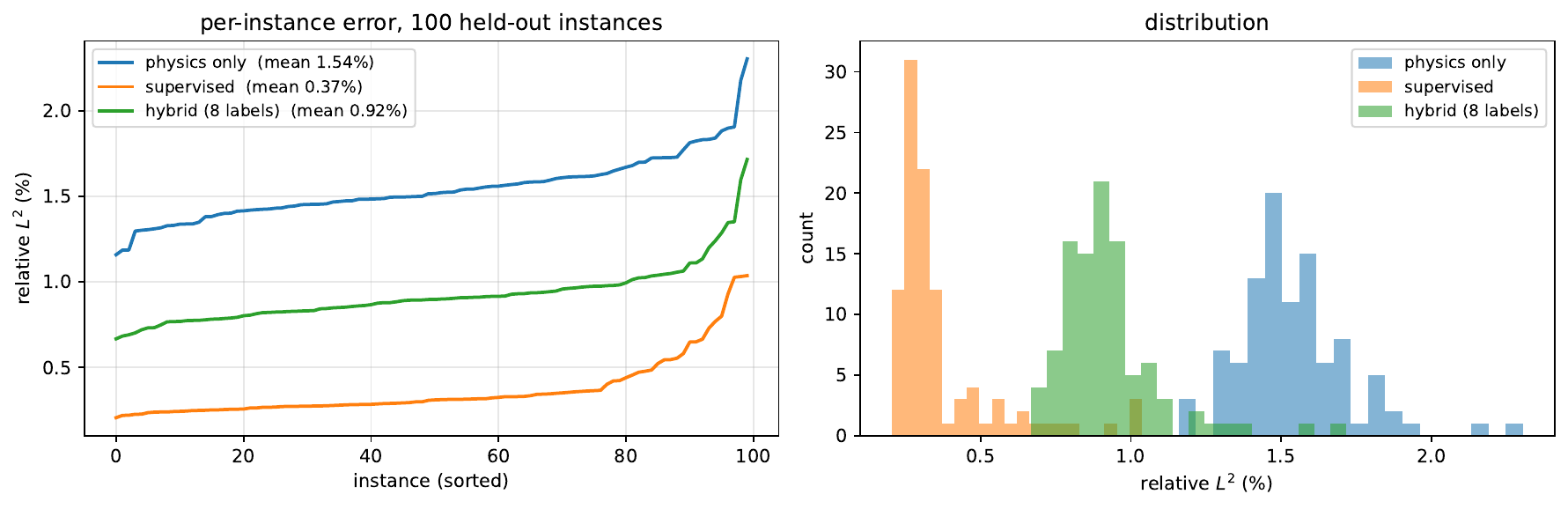}
\caption{Reversed vortex: relative $L^2$ error on each of the $100$ held-out
instances, sorted (left) and as a distribution (right), seed $42$.}
\label{fig:rv-spread}
\end{figure}

\paragraph{Seed variability is small in every arm.}
The standard deviations are $0.067$, $0.081$ and $0.035$ percentage points, an
order of magnitude below the differences between arms. The ranking is therefore
not an artefact of initialisation, and Figure~\ref{fig:rv-spread} shows that it
holds across the test set rather than on average only: the three per-instance
curves do not cross.

\paragraph{The physics term does not confer an area-conservation advantage
here.}
The enclosed-area error follows the same ordering as the field error, with the
supervised arm best at $1.13\%$. This ordering is specific to this benchmark: 
as Table~\ref{tab:eikstat} records, $86.9\%$ of the
domain violates $|\nabla\phi|=1$ in the exact solution, and the adaptive weight
consequently settles at $O(10^{-4})$, and the error fields of
Figure~\ref{fig:rv-fields} show none of the structure that distinguishes the
two arms under rotation. The geometric constraint is therefore
inactive for most of training, and the data-free arm is regularised by the
transport residual alone.

The train and test errors of the data-free arm are nearly equal, $1.503\%$
against $1.543\%$ at seed $42$, so that arm is not limited by the size of the
training family at sixteen instances. The supervised arm separates further,
$0.256\%$ against $0.368\%$, which is the behaviour expected of the arm that
must infer the map from examples.

\begin{figure}[H]
\centering
\includegraphics[width=\textwidth]{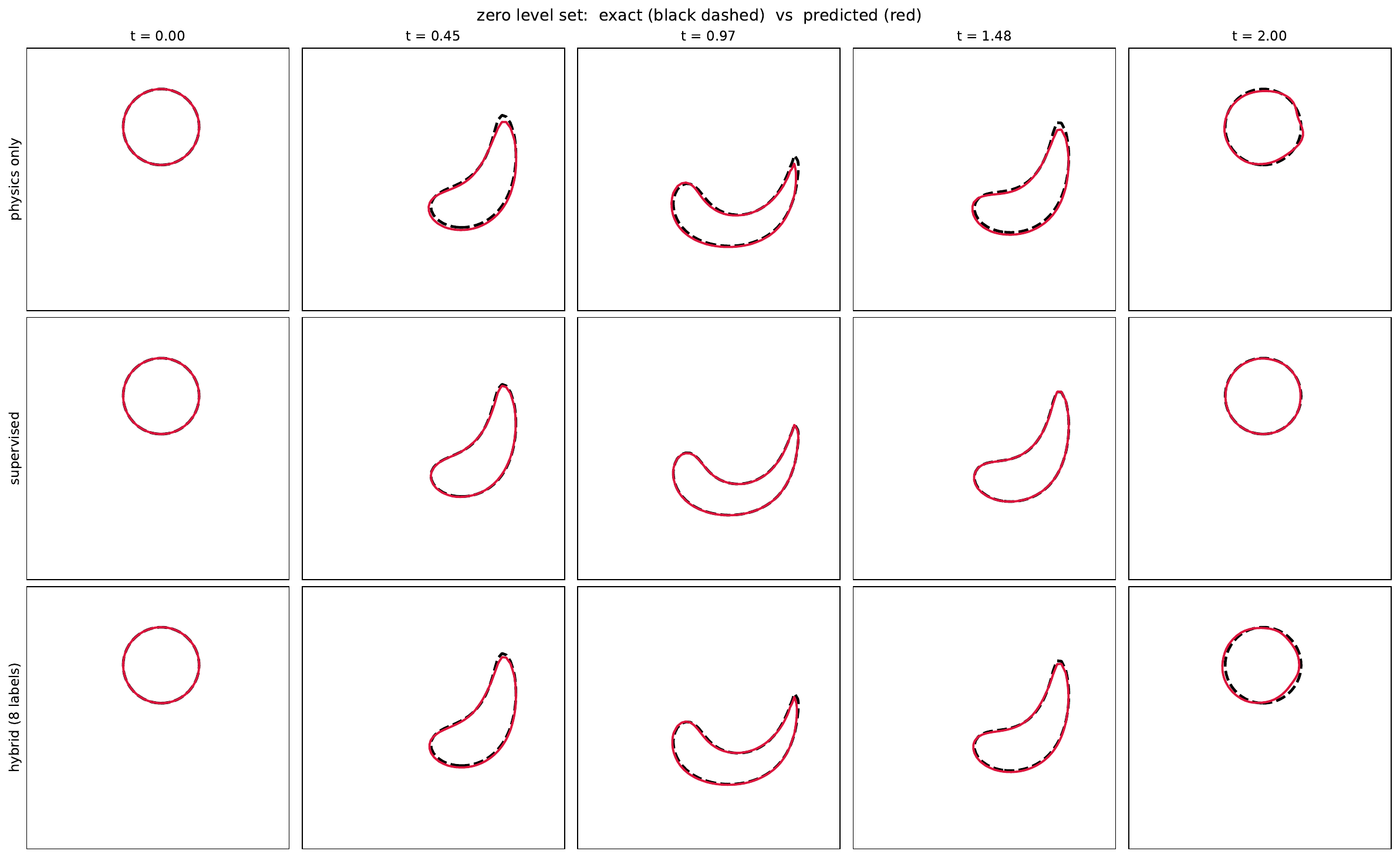}
\caption{Reversed vortex: predicted zero level set (solid) against the exact
interface (dashed) at five times, for one held-out initial interface. Rows are
the three arms at seed $42$.}
\label{fig:rv-interfaces}
\end{figure}

\begin{figure}[H]
\centering
\includegraphics[width=\textwidth]{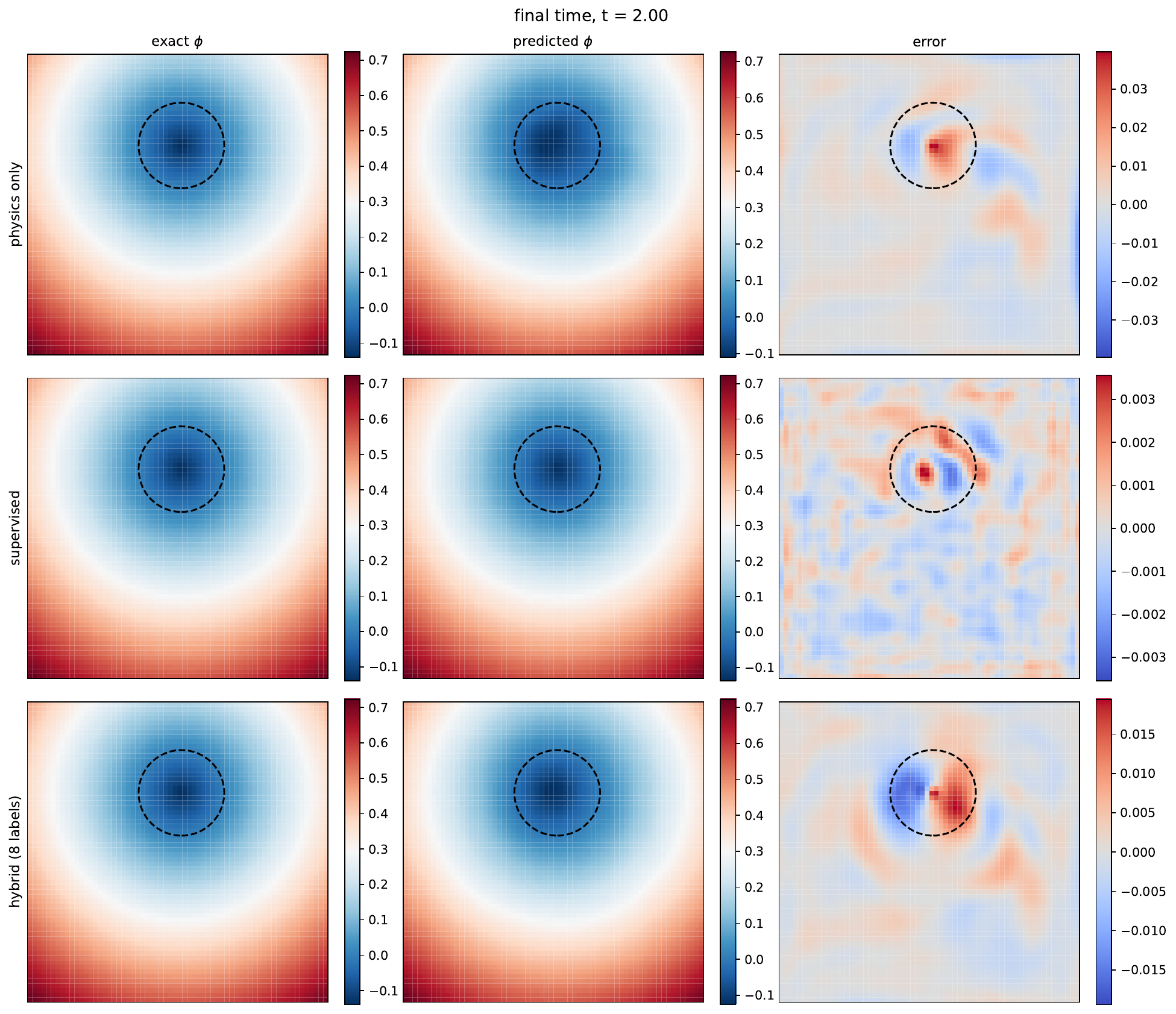}
\caption{Reversed vortex at the final time: exact field, prediction and signed
error for the same held-out instance as Figure~\ref{fig:rv-interfaces}. The
dashed contour marks the exact interface. Note the differing colour scales.}
\label{fig:rv-fields}
\end{figure}

\subsection{The eikonal term and the initial condition}
\label{sec:res:abl}

Two choices in Section~\ref{sec:method} are not forced by the formulation: the
initial condition is imposed structurally rather than by penalty, and the
adaptive eikonal weight is seeded at zero rather than at the first ratio.
Table~\ref{tab:abl} varies each in turn against the data-free arm at seed
$42$, holding everything else fixed, and adds the case in which the eikonal
term is removed altogether.

\begin{table}[H]
\centering
\caption{Data-free arm, seed $42$, with the eikonal treatment and the initial
condition varied one at a time. Sixteen training instances, $100$ held-out
instances; $30{,}000$ iterations for rotation and $20{,}000$ for the vortex.
The final column is the eikonal weight at the end of training.}
\label{tab:abl}
\begin{tabular}{llccc}
\toprule
& \textbf{Configuration} & \textbf{Relative $L^2$ (\%)} &
\textbf{Area error (\%)} & $w_{\text{eik}}$ \\
\midrule
\multirow{4}{*}{\textbf{RO}}
 & zero seeding (as used)      & $4.042$ & $\mathbf{3.15}$ & $1.3\times10^{-3}$ \\
 & ratio seeding (published)   & $\mathbf{3.417}$ & $3.99$ & $2.7\times10^{-3}$ \\
 & soft initial condition      & $6.019$ & $11.50$ & $4.1\times10^{-3}$ \\
 & no eikonal term             & $18.968$ & $5.76$ & --- \\
\midrule
\multirow{4}{*}{\textbf{RV}}
 & zero seeding (as used)      & $1.543$ & $\mathbf{2.67}$ & $6.3\times10^{-5}$ \\
 & ratio seeding (published)   & $24.812$ & $73.64$ & $3.7\times10^{-3}$ \\
 & soft initial condition      & $26.846$ & $91.02$ & $7.7\times10^{-1}$ \\
 & no eikonal term             & $\mathbf{1.266}$ & $3.56$ & --- \\
\bottomrule
\end{tabular}
\end{table}

\paragraph{The eikonal term is essential on one benchmark and a liability on
the other.}
Removing it costs a factor of $4.7$ on rotation and \emph{gains} $18\%$ on the
vortex. The sign of that difference follows Table~\ref{tab:eikstat} directly:
the exact solution of the rotation problem is a signed distance function, so
the constraint carries information about the solution, whereas the vortex
violates it over most of the domain and the constraint carries error instead.
No single fixed weight serves both, which is what the adaptive scheme is for.
The vortex is not best served by discarding the term entirely either: the
weight it selects, $6.3\times10^{-5}$, conserves area
$25\%$ better than a weight of exactly zero, at a cost of $0.28$ percentage
points in field error.

\paragraph{The seeding of the weight decides whether the vortex trains at
all.}
Seeding at the first ratio rather than at zero changes the rotation result by
$15\%$ in one direction and the vortex result by a factor of $16$ in the other.
The asymmetry is the mechanism set out in Section~\ref{sec:eikonal}: a field
that begins at an exact distance function makes the ratio uninformative at the
first iteration, and the weight is then held near its clamp for
$O((1-\beta)^{-1})$ steps. On rotation that costs convergence speed, since the
constraint is valid and a large weight points in roughly the right direction.
On the vortex it destroys the interface, and area conservation degrades from
$2.67\%$ to $73.64\%$.

\paragraph{The soft initial condition is worse on both benchmarks and both
measures.}
Its weight ends at $4.1\times10^{-3}$ on rotation and at $0.77$ on the vortex,
against $1.3\times10^{-3}$ and $6.3\times10^{-5}$ for the structural
constraint. The reason is visible in~\eqref{eq:saw}: an initial-condition loss
adds $\lVert\nabla_\theta\mathcal{L}_{\text{ic}}\rVert$ to the numerator, and
the ratio is inflated for as long as that term is active. On the vortex the
weight never leaves the neighbourhood of the clamp at all, and the area error
reaches $91\%$.

\begin{figure}[H]
\centering
\includegraphics[width=\textwidth]{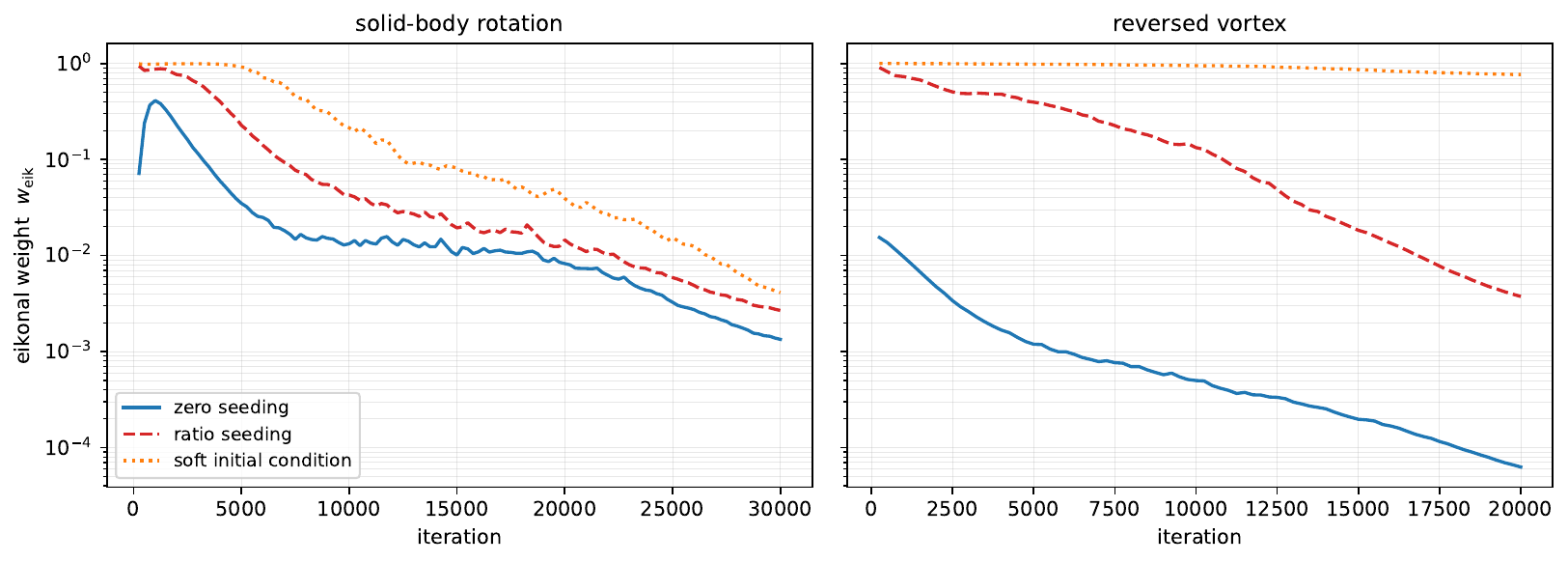}
\caption{Eikonal weight against training iteration for the three
configurations of Table~\ref{tab:abl} that carry the eikonal term, seed $42$. Note the logarithmic
ordinate. The soft initial condition holds the weight within a factor of two
of its clamp for the whole of the vortex run.}
\label{fig:weights}
\end{figure}

Figure~\ref{fig:weights} shows the three weight trajectories. Two features are
not apparent from the final values alone. Zero seeding does not decay
monotonically: on rotation the weight rises from $0.07$ to a maximum near
$0.4$ within the first thousand iterations before descending, which is the
ratio becoming informative as the field departs from $\phi_0$. And the three
trajectories converge to within a factor of three by the end of the rotation
run, whereas on the vortex they remain two orders of magnitude apart
throughout. The seeding is therefore a matter of convergence rate on one
benchmark and of whether the interface survives at all on the other.

\subsection{Sensitivity to the optimisation budget}
\label{sec:res:budget}

The rotation benchmark was run at both $20{,}000$ and $30{,}000$ iterations
with everything else held fixed. Table~\ref{tab:budget} reports both. Every arm
improves on both measures at the larger budget, so the shorter run is not
converged; the improvement is largest for the data-free arm, which is
consistent with the slower convergence of a two-term objective. The seed spread
of that arm, however, narrows only from $1.277$ to $1.075$ percentage points,
which is the basis for attributing it to the adaptive weight rather than to
truncation.

\begin{table}[H]
\centering
\caption{Solid-body rotation at two optimisation budgets, all other settings
identical. Three seeds, $100$ held-out instances.}
\label{tab:budget}
\begin{tabular}{lcccc}
\toprule
& \multicolumn{2}{c}{\textbf{Relative $L^2$ (\%)}}
& \multicolumn{2}{c}{\textbf{Area error (\%)}} \\
\cmidrule(lr){2-3}\cmidrule(lr){4-5}
\textbf{Arm} & $20{,}000$ & $30{,}000$ & $20{,}000$ & $30{,}000$ \\
\midrule
Data-free  & $4.716 \pm 1.277$ & $3.804 \pm 1.075$ & $4.77 \pm 0.74$ & $3.15 \pm 0.49$ \\
Hybrid     & $1.792 \pm 0.119$ & $1.599 \pm 0.084$ & $3.84 \pm 0.40$ & $3.42 \pm 0.28$ \\
Supervised & $2.621 \pm 0.185$ & $2.576 \pm 0.159$ & $9.13 \pm 1.30$ & $8.61 \pm 0.97$ \\
\bottomrule
\end{tabular}
\end{table}

The vortex benchmark reaches the same values at $20{,}000$ iterations as at
larger budgets and is reported at that figure. The difference is consistent
with the eikonal weight: it is suppressed to $O(10^{-4})$ on the vortex,
leaving a single-objective loss, and remains active on rotation throughout.

\section{Discussion}
\label{sec:discussion}

\subsection{What the eikonal constraint explains}

The two benchmarks rank the three arms differently, and the ranking is not a
matter of one being harder than the other. Under the vortex the ordering
follows the label count monotonically; under rotation it does not, the hybrid
arm outperforming the supervised arm with half as many reference solutions, and
the data-free arm conserving enclosed area best of the three despite the
largest field error.

The measurement in Table~\ref{tab:eikstat} separates the two cases. Rotation is
rigid, so the exact solution remains a signed distance function and the
constraint $|\nabla\phi|=1$ is satisfied by the quantity being approximated. It
therefore supplies genuine information about the solution, and the arms that
carry it inherit that information. The vortex deforms the field, the exact
solution violates the constraint over $86.9\%$ of the domain, and the adaptive
weight suppresses the term to $O(10^{-4})$; the physics-trained arms are then
regularised by the transport residual alone and the advantage disappears.

This is a statement about the level-set formulation rather than about neural
operators. The signed-distance property is a convenience of the representation,
not a property of the solution, and it is preserved only under rigid motion.
Classical schemes address the same tension by reinitialising between transport
steps~\citep{Sussman1994,SussmanFatemi1999}, which restores the condition
without moving the zero contour. A trained operator has no equivalent
intermediate step, so the constraint enters the objective and competes with the
residual for the whole of training. Where it is valid that competition is
productive; where it is not, no choice of weight recovers the lost accuracy,
and the adaptive scheme correctly turns the term off.

\subsection{Field error and interface error are not the same quantity}

On rotation the supervised arm attains the second-best relative $L^2$ error and
the worst enclosed-area error by a factor of $2.7$. The two measures therefore
order the arms differently, and reporting either alone would misrepresent the
comparison.

Figure~\ref{fig:ro-fields} shows the mechanism. The supervised residual is
high-frequency and spread over the whole domain, including the band where the
zero contour lies; the data-free residual is smooth, largest away from the
interface, and organised along the medial axis where the distance function
fails to be differentiable. A mean-squared error against solution data
penalises deviation uniformly and has no term that acts specifically on the
interface, whereas the transport residual constrains the whole field to evolve
consistently and the geometric term acts precisely where the gradient magnitude
matters.

For level-set problems the enclosed volume is the quantity that classical
schemes are judged on, and the same standard should apply here.

\subsection{Cost, and where data-free training is worth its cost}

The benchmarks used here admit exact solutions, in closed form for rotation and
by characteristic tracing for the vortex, so reference data are inexpensive and the supervised baseline is not disadvantaged by the cost of
obtaining them. That is a weakness of the motivation and a strength of the
study: it is what permits the price of refusing labels to be measured rather
than asserted, and the measured price is a factor of $1.5$ on rotation and
$4.4$ on the vortex.

The regime in which that price is worth paying is the one where the velocity
field is itself a solution variable. Characteristics do not exist when
$\mathbf{u}$ is obtained from a momentum balance coupled to $\phi$, every
training label then requires a full two-phase solve, and the comparison made
here would be dominated by the cost of generating the supervised arm's training
set rather than by its accuracy. WINO~\citep{zhu2026wino} reports exactly this
balance for a problem where each reference costs a nonlinear solve: the
data-free variant is faster overall by a factor of $4.5$, almost all of which
is reference generation rather than training.

A second consideration is the training set itself. Instances here are initial
interfaces, which are prescribed data in any level-set computation and cost
nothing to generate; the data-free arm is therefore limited by optimisation
rather than by the availability of training inputs. Its train and test errors
differ by less than a factor of $1.1$ on both benchmarks.

\subsection{Limitations}

Three limitations bear on how the results should be read.

The velocity field is held fixed within each benchmark. The operator is
therefore over initial interfaces, which is the protocol used throughout the
operator literature, but the broader map over $(\phi_0,\mathbf{u})$ is not
demonstrated.

The data-free arm on rotation exhibits a seed spread an order of magnitude
larger than the other two arms, traced in Section~\ref{sec:res:ro:a} to the
value at which the adaptive eikonal weight settles. This does not affect the
ranking, whose separation is far larger than the spread, but it is a property
of the adaptive scheme on benchmarks where the constraint is active and it is
not removed by a longer budget.

The benchmarks are two-dimensional, use smooth circular interfaces, and are
resolved on a $64\times64$ grid.

\section{Conclusion}
\label{sec:conclusion}

This work has developed a neural operator for level-set interface advection
trained without reference solutions. The operator maps an initial interface to
the full spatiotemporal trajectory under a prescribed flow in a single forward
pass, and its parameters are determined entirely from the transport residual
and a geometric constraint. Supervised and hybrid operators trained under
identical conditions quantify what the absence of labels costs.

On a reversed single vortex the data-free operator attains
$1.614 \pm 0.067\%$ relative $L^2$ error on $100$ held-out initial interfaces,
against $0.369 \pm 0.035\%$ for a supervised operator trained on sixteen exact
solutions. On solid-body rotation the figures are $3.804 \pm 1.075\%$ and
$2.576 \pm 0.159\%$. The cost of refusing labels is therefore a factor of $4.4$
in the first case and $1.5$ in the second, and the difference between them is
accounted for by the eikonal constraint, which is satisfied by the exact
solution under rigid motion and violated over most of the domain under
deformation. Where the constraint is valid, physics-based training conserves
enclosed area substantially better than supervision on solution data, and eight
reference solutions combined with the residual outperform sixteen without it.

Two directions follow. The natural extension is an operator over the velocity
field as well as the initial interface, which the present formulation admits
without structural change. The more consequential one is the coupled setting,
in which the velocity is obtained from a momentum balance rather than
prescribed: characteristic tracing is then unavailable, every reference
solution requires a full two-phase computation, and the comparison reported
here would be decided by the cost of the training set rather than by the
accuracy gap measured on it.

\section*{Acknowledgments}
The author acknowledges the computational resources provided by NED University
of Engineering and Technology, Karachi, and by Google Colab.

\section*{Funding}
This research received no specific grant from any funding agency in the public,
commercial, or not-for-profit sectors.

\section*{Conflict of Interest}
The author has no conflicts to disclose.

\section*{Author Contributions}
\textbf{Muhammad Akbar Khan}: Conceptualization, Methodology,
Software, Formal analysis, Data curation, Visualization,
Writing -- original draft, Writing -- review \& editing.

\section*{Data and Code Availability}
The implementation, the notebooks reproducing every experiment reported here,
and the run records from which every table and figure is generated are
available at~\citep{khan2026dfno_code}.

\bibliographystyle{plainnat}
\bibliography{references}

\end{document}